\documentclass[letterpaper]{article} 
\usepackage{aaai25}  
\usepackage{times}  
\usepackage{helvet}  
\usepackage{courier}  
\usepackage[hyphens]{url}  
\usepackage{graphicx} 
\usepackage{natbib}  
\usepackage{caption} 
\usepackage{algorithm}
\usepackage{algorithmic}
\usepackage{booktabs}
\usepackage{multirow}
\usepackage{amsmath}
\usepackage{amssymb}
\usepackage{subfig}
\usepackage{xcolor}

\RequirePackage{xspace}
\makeatletter
\DeclareRobustCommand\onedot{\futurelet\@let@token\@onedot}
\def\@onedot{\ifx\@let@token.\else.\null\fi\xspace}

\def\ie{\emph{i.e}\onedot}

\def\etal{\emph{et al}\onedot}

\newcommand{\answerTODO}[1][]{\textcolor{red}{\bf [TODO]}}
\newcommand{\justificationTODO}[1][]{\textcolor{red}{\bf [TODO]}}

\newsavebox\CBox
\def\textBF#1{\sbox\CBox{#1}\resizebox{\wd\CBox}{\ht\CBox}{\textbf{#1}}}
\newtheorem{canonical}{Definition}
\usepackage{newfloat}
\usepackage{listings}
\DeclareCaptionStyle{ruled}{labelfont=normalfont,labelsep=colon,strut=off} 
\floatstyle{ruled}
\newfloat{listing}{tb}{lst}{}
\floatname{listing}{Listing}

\title{Deep Non-rigid Structure-from-Motion Revisited: \\Canonicalization and Sequence Modeling}
\author{
    Hui Deng\textsuperscript{\rm 1}, 
    Jiawei Shi\textsuperscript{\rm 1}, 
    Zhen Qin\textsuperscript{\rm 2}, 
    Yiran Zhong\textsuperscript{\rm 3}, 
    Yuchao Dai\textsuperscript{\rm 1}\footnote{ Corresponding author}
}
\affiliations{
    \textsuperscript{\rm 1} School of Electronics and Information, Northwestern Polytechnical University\\
    \textsuperscript{\rm 2} TapTap \\
    \textsuperscript{\rm 3} Shanghai AI Lab \\

    \{denghui986, sjw2018\}@mail.nwpu.edu.cn, \{zhenqin950102, zhongyiran\}@gmail.com, daiyuchao@nwpu.edu.cn
}

\usepackage{bibentry}

\begin{document}

\maketitle

\begin{abstract}

Non-Rigid Structure-from-Motion (\textbf{NRSfM}) is a classic 3D vision problem, where a 2D sequence is taken as input to estimate the corresponding 3D sequence. Recently, deep neural networks have greatly advanced the task of NRSfM. However, existing deep NRSfM methods still have limitations in handling the inherent sequence property and motion ambiguity associated with the NRSfM problem. In this paper, we revisit deep NRSfM from two perspectives to address the limitations of current deep NRSfM methods: \textbf{(1)} canonicalization and \textbf{(2)} sequence modeling.
We propose an easy-to-implement per-sequence canonicalization method as opposed to the previous per-dataset canonicalization approaches. With this in mind, we propose a sequence modeling method that combines temporal information and subspace constraints. As a result, we have achieved a more optimal NRSfM reconstruction pipeline compared to previous efforts. The effectiveness of our method is verified by testing the sequence-to-sequence deep NRSfM pipeline with corresponding regularization modules on several commonly used datasets.

\end{abstract}

\vspace{-2mm}
\section{Introduction}
\label{Introduction}

Non-Rigid Structure-from-Motion (NRSfM) is one of the classical tasks in 3D computer vision, which aims at recovering the 3D deforming shape sequences from 2D observation sequences.
There has been a lot of work trying to construct a reasonable model for this task. Various traditional mathematical methods~\cite{akhter2008nonrigid, akhter2009defense, dai2014simple,zhu2014complex} based on the factorization framework\cite{tomasi1992shape}, represented by Bregler~\etal~\cite{bregler2000recovering}, have produced remarkable results in modeling the deformation sequences.

Along with the rise of machine learning techniques and the advantages shown by neural network models in terms of inference speed and computational accuracy, more people have started to combine deep learning with non-rigid 3D reconstruction~\cite{novotny2019c3dpo, kong2020deep, wang2020deep, zeng2022mhr, wang2021paul, zeng2021pr, park2020procrustean}, and some progress has been made. These methods have certain advantages over traditional methods in terms of computational accuracy, inference speed, and handling of shape ambiguity. However, they also encountered the same problem as traditional methods, namely motion ambiguity, where ambiguity interferes with the direction of convergence of the network, which in turn leads to poor estimation results. To cope with this problem, C3dpo\cite{novotny2019c3dpo} proposes the concept of transversal property and designs the canonicalization network to suppress the ambiguity throughout the data set. On the other hand, PRN\cite{park2020procrustean} represents a sequential regularization method based on Procrustean Analysis(PA) and an elegant training method that eliminates the need for the network to perform complex PA during inference. However, the implementation and computational complexity of this network in training are relatively high. At the same time, the backbone network of these methods\cite{novotny2019c3dpo,park2020procrustean,sidhu2020neural,zeng2021pr} only performs lifting on 2D observations and lacks modeling of continuous 2D observation sequences.


Recently, Deng~\etal\cite{deng2022deep}({denoted as Seq2Seq}) designs sequence a reconstructing module
by referring to the traditional mathematical modeling of sequences, and has achieved promising results. However, instead of tackling canonicalization in a sequential manner, this method follows \cite{novotny2019c3dpo} and deals with the problem throughout the dataset. In addition, this method models the sequence with little consideration of the positional information in the sequence structure, and only an absolute positional encoding bias is used to inject temporal information into the sequence reconstruction process.

In this paper, we propose to revisit the deep NRSfM problem. We argue that dealing with shape ambiguity in a \emph{sequence-by-sequence} manner is more effective than over the entire dataset, reducing the effect of ambiguity on the estimation results. We propose a sequential canonicalization based on General Procrustean Analysis(GPA). This method focuses on a parameter-free GPA layer, which avoids high training overhead and eliminates the need for complex GPA operations during inference.
In conjunction with the canonicalization for sequences, we design a sequence modeling method that can utilize sequence temporal more effectively. We argue that using temporal bias alone to introduce temporal information for sequence reconstruction is insufficient, because there is no guarantee that the network will understand what the bias represents.

Our main contributions are summarized as follows:
\begin{itemize}
\item A sequence-by-sequence canonicalization method with an easy-to-implement GPA Layer, effectively reducing the impact of the inherent ambiguity of NRSfM on estimation accuracy;
\item A method that can efficiently incorporate temporal information to reconstruct non-rigid sequences for better shape sequence estimation results; 
\item We conducted experiments on a substantial amount of data, and the results confirm the validity of our method in most scenarios.
\end{itemize}

\section{Related Work}

\subsection{Classical NRSfM Methods}
Bregler~\etal~\cite{bregler2000recovering} first introduced factorization framework~\cite{tomasi1992shape} for Non-rigid Structure-from-Motion.
Xiao~\etal~\cite{xiao2004closed} pointed out that the recovery of shape basis has inherent ambiguity problems. Akhter~\etal~\cite{akhter2009defense} showed that the non-rigid 3D shapes can be recovered uniquely by suitable optimization method under the orthogonal constraint. Subsequent researchers have proposed many effective NRSfM methods, including using DCT basis to recover motion trajectory\cite{akhter2008nonrigid, gotardo2011computing}, using low-rank constraint to regularize 3D structure recovery\cite{paladini2010sequential, dai2014simple, kumar2020non, kumar2022organic}, using hierarchical prior\cite{torresani2008nonrigid}, using union-of-subspace constraint\cite{zhu2014complex, kumar2016multi, kumar2017spatio}, using procrustean alignment to refine camera motion\cite{lee2013procrustean, Lee_2014_CVPR, park2017procrustean}, using consensus prior\cite{cha2019reconstruct}, using manifold to model non-rigid 3D surface\cite{kumar2018scalable, parashar2017, parashar2020local}, and many other quintessential methods\cite{marques2008optimal, paladini2009factorization, garg2013dense, simon2016kronecker, agudo2016recovering}.


\subsection{Deep NRSfM Methods}
Novotny \etal \cite{novotny2019c3dpo} proposed the first deep learning framework for NRSfM. They design a decomposition network to predict shape and motion separately, and use canonical loss to fix the estimated 3D shapes in the canonical coordinate. Kong \etal \cite{kong2020deep} used the sparse representation assumption to build a hierarchical network for non-rigid 3D reconstruction. Wang \etal \cite{wang2020deep} extended this method to the perspective projection model. Park \etal \cite{park2020procrustean} extended the alignment constraint in Procrustean Regression\cite{park2017procrustean} to the deep learning method, avoiding the complex camera motion estimation problem. Zeng \etal \cite{zeng2021pr} further strengthened the inter-frame structure connection by introducing a regularization constraint for the pairwise alignment of 3D shapes. Wang \etal \cite{wang2021paul} 
propose to use the orthographic-N-point algorithm to regularize auto-encoder to obtain better reconstruction.  Sidhu~\etal~\cite{sidhu2020neural} designs the trajectory energy function for imposing subspace constraints, accompanied by a regular term for temporal smoothing.  However, the method does not combine subspace constraints with sequential temporal constraints
Recently, Zeng \etal \cite{zeng2022mhr} focused on the depth ambiguity problem and modeled the generation of multi-hypothesis through the introduction of Gaussian noise.

Deep Sparse NRSfM Methods often directly use the network to regress the coordinates of 3D keypoints corresponding to 2D annotations, while dense NRSfM tasks often use mesh or implicit neural representation to represent 3D structures. Yang \etal \cite{yang2021lasr, yang2021viser, Yang_2022_CVPR} proposed a framework that establishes the correspondence between long-sequence observation points by optical flow and model the deformation with skin model. In this paper, we focus on the reconstruction of sparse non-rigid 3D structures.

\vspace{-1mm}

\subsection{Sequence Modeling}
The introduction of neural networks in sequence processing problems can improve processing efficiency and has been widely used in many sequence-to-sequence processing tasks\cite{hinton2012deep, krizhevsky2017imagenet, he2017mask}. Recently, the proposal of Transformer \cite{vaswani2017attention} has greatly promoted the development of the sequence modeling network. Transformer and its varieties 
has been extended to many other fields, such as human body pose estimation\cite{kocabas2020vibe, zhang2022mixste}, action recognition\cite{choi2019can}, NRSfM\cite{deng2022deep}, \emph{etc}.

Some methods do not adopt the Transformer structure, but try to deal with sequence problems from other perspectives. Wu \etal \cite{wu2022timesnet} used the time series analysis method to calculate the spectrum information of the sequence.
Qin \etal \cite{qin2023toeplitz} proposed to use the relative position encoded Toeplitz matrix to replace the attention to realize the function of the tokens mixer. In deep non-rigid 3D reconstruction tasks, some methods strengthen the network's understanding of sequences. Park \etal \cite{park2020procrustean} used the Procrustean alignment to design an optimization equation and extended it to a loss function.
Zeng \etal \cite{zeng2021pr} constructed the internal constraints of the 3D structures through pairwise contrastive and consistent regularization.

\vspace{-2mm}

\section{Method}

\subsection{Preliminary}
\begin{figure*}[t]
  \centering
  \includegraphics[width=0.8\linewidth]{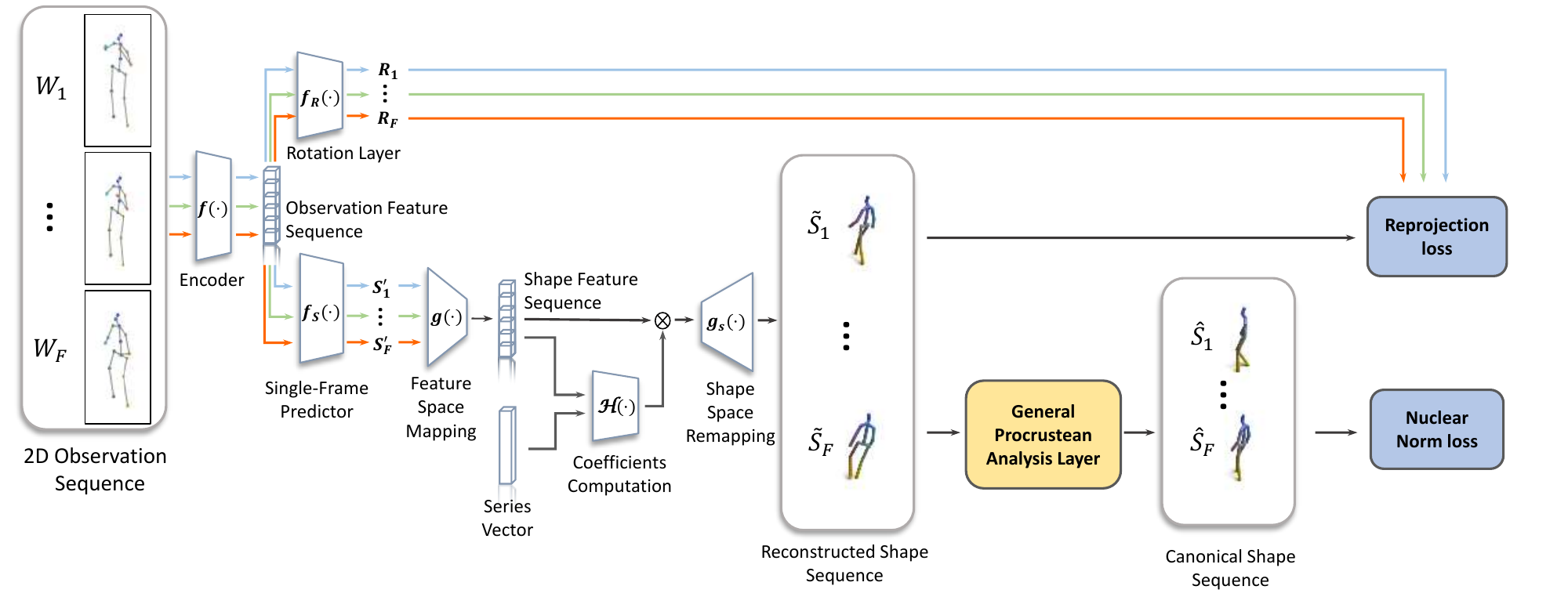}
  \caption{An overview of deep NRSfM pipeline with proposed shape sequence reconstruction and GPA layer. The whole pipeline consists of three parts: the single-frame shape/rotation predictor, the shape sequence reconstruction stage and general procrustean analysis layer. The General Procrustean Analysis layer is a parameter-free layer, which is not needed for inference when training is finished. }
  \label{fig:pipeline}

\end{figure*}

We first introduce the definition of this problem and review some classical solutions to provide possible thinking paths. The following equation gives the general form of NRSfM:
\begin{equation}
\begin{aligned}
\label{eq:motion_shape}
\mathbf{W}
&= \begin{bmatrix}
\mathbf{\Pi}_{1}  &  & \\ 
 & \ddots  & \\ 
 &  & \mathbf{\Pi}_{F} 
\end{bmatrix} \left[
    \begin{array}{ccc}
      \mathbf{R}_1 &  &  \\
       & \ddots &  \\
       &  & \mathbf{R}_F \\
    \end{array}
  \right]\left[
           \begin{array}{c}
             \mathbf{S}_1 \\
             \vdots \\
             \mathbf{S}_F \\
           \end{array}
         \right]
         \\ &= \mathbf{\Pi}\mathbf{R} \mathbf{S}, \mathbf{W}\in\mathbb{R}^{2F\times P},\mathbf{R}\in\mathbb{R}^{3F\times3F},\mathbf{S}\in\mathbb{R}^{3F\times P},
\end{aligned}
\end{equation}
where $\mathbf{\Pi}_{i}=\begin{bmatrix}
1 & 0 & 0;0 & 1 & 0
\end{bmatrix}$ is the orthogonal projection matrix, $\mathbf{W}$ is the matrix composed of 2D observations of continuously deforming objects, $\mathbf{S}$ is the 3D shape matrix corresponding to the 2D observations, and $\mathbf{R}$ is the motion matrix. 
It should be noted that in the above modeling, no distinction is made between the ego-motion of the camera and the motion of the object:
\begin{equation}
    \label{eq:motion_ambiguity}
\mathbf{W}=\mathbf{\Pi}\mathbf{R}_{motion}\mathbf{S}=\mathbf{\Pi}\mathbf{R}_{camera}\mathbf{R}_{shape}\mathbf{S}.
\end{equation}

The underdetermination defect caused by this ambiguity of coupled motion has resulted in the loss of reconstruction accuracy in traditional methods and deep methods, and how to effectively deal with this problem has become one of the focuses of discussion in various works.

The solution model defined by Eq.~\eqref{eq:motion_shape} is usually transformed into an optimization model with additional constraints, and one of the most representative work\cite{dai2014simple} gives the optimization condition by analyzing the rank of the sequence matrix to suppress ambiguity:
\begin{equation}
    \min_{\mathbf{S}}{\|\mathbf{S}^\sharp\|_*}, s.t. \mathbf{W}=\mathbf{\Pi}\mathbf{RS},\mathbf{S}\in\mathbb{R}^{3F\times P},\mathbf{S}^\sharp\in\mathbb{R}^{F\times3P},
\end{equation}
where $\mathbf S^\sharp$ is a reshuffled form of $\mathbf S$ as described in Dai~\etal~\cite{dai2014simple}. The most important feature of this model is that no additional priori assumptions are introduced for the solution, but rather a low-rank constraint.
Later, Zhu~\etal~\cite{zhu2014complex} noted on the basis of Dai~\etal\cite{dai2014simple} that this low-rank constraint on deforming sequences can be extended to a union-of-subspace representation. This work assumes that deforming shapes in the same sequence should come from one or more intersecting shape spaces, each of which is spanned by a set of shape basis:
\begin{equation}
    \min_{\mathbf{D}}{\|\mathbf{D}\|_*}, s.t. \mathbf{W} = \mathbf{\Pi}\mathbf{R}\mathbf{S}, \mathbf{S} = \mathbf{D}\mathbf{C},
        \label{eq:union_subspace}
\end{equation}
where $\mathbf{D}$ is a dictionary matrix containing multiple bases that can be tensed into different subspace shapes, and $\mathbf{C}$ is a coefficient matrix. This sequence modeling method referred to LRR\cite{liu2012robust} brings new effective constraints to the solution process, refining the regularity of the low-rank constraint on the estimation results.


However, these modeling of sequences are rarely discussed in subsequent deep methods, and the mainstream deep methods~\cite{novotny2019c3dpo,kong2020deep,wang2021paul} reconstruct the non-rigid shape sequences in a \emph{single-frame lifting} way as follows:
\begin{align}
    \label{eq:single_net}
    &\mathbf{S'}_i = f_{S}(f(\mathbf{W}_i),\Theta_{\mathbf S}), \mathbf{S'}_i \in \mathbb{R}^{3\times P},\\
    &\mathbf{R}_i = f_{R}(f(\mathbf{W}_i),\Theta_{\mathbf R}), \mathbf{R}_i \in \mathbb{R}^{3\times3},
\end{align}
where $\Theta$ is learnable parameters, and $f:\mathbb{R}^{3\times P}\to\mathbb{R}^{D}$ is an encoder module. Whilst such methods heavily rely on a network's ability to recall the relevant 3D structure from a sole frame of data, it is evidently apparent that they are incapable of utilizing constraints from \emph{sequence context, temporal information, and the sequence structure itself} which can be provided to facilitate the reconstruction process.

In this paper, we revisit the canonicalization and the sequence structure self-expressive regularity under sequence modeling, and design modules that can effectively impose both constraints to obtain a deeply non-rigid reconstruction pipeline that can effectively exploit the sequence structure information. The entire pipeline is shown in Fig.~\ref{fig:pipeline}, and we will then present first how to perform canonicalization in Sec.~\ref{sec:GPA}, followed by sequence modeling in Sec.~\ref{sec:seq_recon}.

\subsection{General Procrustean Analysis Layer}
\label{sec:GPA}

To reduce the ambiguity described as Eq.~\eqref{eq:motion_ambiguity} for alleviating the loss of accuracy on estimating shapes, we design a computational module $\mathcal{G}: \mathbb{R}^{F\times 3P}\to\mathbb{R}^{F\times 3P}$ that constrains the sequence estimated by the network to the canonical coordinate system originally proposed by C3DPO~\cite{novotny2019c3dpo}.
               
\begin{canonical}
    \label{def:canoincal}
    A deforming shape set $\mathcal{S}=\{\mathbf{S}_1,\cdots,\mathbf{S}_n\}$ is in canonical coordinate, $\forall \mathbf{S,S'}\in\mathcal{S}$ and $\mathbf{R}\in\text{SO}(3)$ $\mathbf{S}=\mathbf{R}\mathbf{S'}$ i.f.f. $\mathbf{S} = \mathbf{S'},\mathbf{R}=\mathbf{I}_3$.
\end{canonical}
Unlike C3dpo, we define the input sequence rather than the entire dataset as a deforming shape set $\mathcal{S}$.


Thus, the previously coupled motion decomposition problem is treated as solving only one motion $\mathbf R_{motion}$. It should be noted that the motion $\mathbf R_{motion}$ here is actually no longer the camera motion but a combination of the origin camera motion and shape rigid motion, $\mathbf R_{motion}= \mathbf R_{camera}\mathbf R_{shape}$, since NRSfM focuses more on the recovery of the nonrigid shapes.


\begin{figure*}[t]
  \centering
  \includegraphics[width=0.8\linewidth]{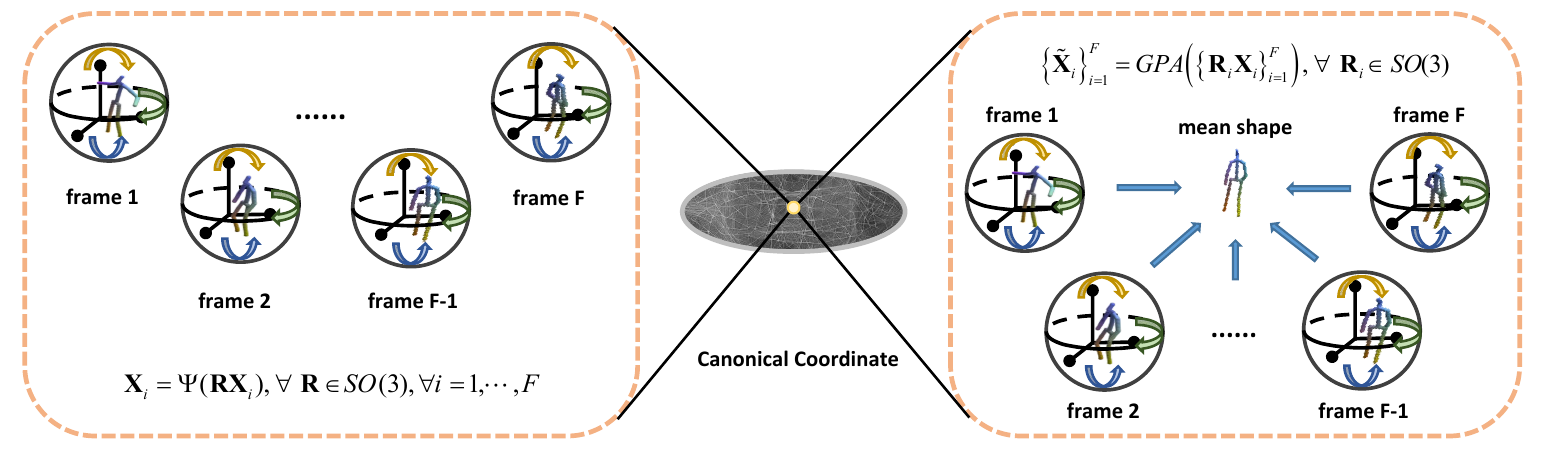}
  \caption{The left side of the figure shows the canonicalization method in \cite{novotny2019c3dpo}, which performs random rotations for each frame, \ie, training the network over the \emph{entire dataset} to determine the canonical coordinate. The right side shows our method to align for \emph{each sequence}, explicitly helping the network to determine the canonical coordinate.}
  \label{fig:canonical}
\end{figure*}

In contrast to another Deep NRSfM classical work C3dpo\cite{novotny2019c3dpo} Seq2Seq~\cite{deng2022deep} differs in that it does not need random sampling in the rotation space $\text{SO}(3)$ for each frame of 3D shape and training the network to be invariant to arbitrary rotational transformations. The context layer in Seq2Seq grants greater focus on estimating the sequence structure. The specific principle is shown in Fig.~\ref{fig:canonical}.

Inspired by PND~\cite{lee2013procrustean}, PR~\cite{park2017procrustean} and PRN~\cite{park2020procrustean}, we design a parameter-free GPA Layer, which uses Generalized Procrustean Analysis~\cite{gower1975generalized} to model the canonical coordinate and employs an iterative approach to obtain aligned shapes. 
The goal of the GPA layer is to eliminate rigid motion between any two frames as much as possible in the least squares sense.
The GPA layer takes 3D sequence $\mathbf{\tilde{S}}$ as input and outputs the aligned 3D shapes $\mathbf{\hat{S}}$, which is defined as follows:
\begin{equation}
    \mathbf{\hat{S}} = \mathbf{\hat{R}}\mathbf{\tilde{S}},~ \mathrm{and}~ \left \{ \mathbf{\hat{R}}_{i} \right \}_{i=1}^{F}=\arg \min_{\mathbf{R}_{i}}\sum_{i=1}^{F}\left \| \mathbf{R}_{i}\mathbf{\tilde{S}}_{i} - \mathbf{\bar{S}}\right \|_{F}^{2},
\end{equation}
where $\mathbf{\bar{S}}$ is the mean shape of aligned shapes, which is updated during the iteration, and $\mathbf{\tilde{S}}$ is the reconstructed 3D shape which will be introduced in Sec.~Sequence Reconstruction.
The aligned shapes $\mathbf{\hat{S}}$ are under a fixed coordinate and can naturally be added with some regularization, such as low-rank \emph{etc}.

PRN \cite{park2020procrustean} gives an elegant derivation of how to introduce constraints using GPA in gradient backpropagation, but is accompanied by a high computational overhead. Here we have an approach that is simpler in terms of implementation, but has a similar level of effectiveness. We refer to the iterative gradient backpropagation procedure of \cite{zheng2021ham} and use the one-step gradient strategy to approximate the gradient of the GPA layer during the iteration.

Next, we show how the gradient of this computational procedure should be propagated. Suppose that the regularization term constraint acting on $\mathbf{\hat{S}}$ is $\mathcal{G}(\cdot)$, then the gradient can be calculated according to the chain rule similar to \cite{park2020procrustean}:
\begin{equation}
   \frac{\partial \mathcal{G}}{\partial \mathbf{S}} = \left \langle \frac{\partial \mathcal{G}}{\partial \mathbf{\hat{S}}}, \frac{\partial  \mathbf{\hat{S}}}{\partial \mathbf{S}} \right \rangle,
\end{equation}
where $\left \langle \cdot, \cdot   \right \rangle$ represents the inner product. Once $\mathcal{G}$ is defined, $\frac{\partial \mathcal{G}}{\partial \mathbf{\hat{S}}}$ can be directly derived. Since computing $\mathbf{S}\!\rightarrow\!\mathbf{\hat{S}}$ is an iterative process with an unknown number of iterations, we represent $\frac{\partial \mathbf{\hat{S}}}{\partial \mathbf{S}}$ into iterative form according to the chain rule:
\begin{equation}
    \frac{\partial \mathbf{\hat{S}}}{\partial \mathbf{S}} = \left \langle \frac{\partial \mathbf{\hat{S}}}{\partial \mathbf{S}^{k}},\left \langle \frac{\partial \mathbf{{S}}^{k-1}}{\partial \mathbf{S}^{k-2}}, \left \langle \cdots, \left \langle \frac{\partial \mathbf{{S}}^{2}}{\partial \mathbf{S}^{1}}, \frac{\partial \mathbf{{S}}^{1}}{\partial \mathbf{S}}\right \rangle \cdots \right \rangle   \right \rangle \right \rangle
\end{equation}
where $\mathbf{S}^{k}$ is the intermediate state in the iterative process. This back-propagation process allows the shape sequence $\tilde{\mathbf{S}}$ estimated by our backbone network to be in the canonical coordinate as much as possible.

\subsection{Sequence Reconstruction}
\label{sec:seq_recon}


In the previous section, we introduce the GPA layer, which aims to reduce motion ambiguity by aligning the shape sequences to the canonical coordinate, and in this section, we describe what properties should exist for the sequence. Inspired by Eq.~\eqref{eq:union_subspace}, 
sequence can be represented as follows:
\begin{equation}
    \mathbf{S} = \mathbf{C}\mathbf{S}, \mathbf{S}=[\mathbf{S}^\prime_1,\cdots,\mathbf S^\prime_F]^\intercal, \mathbf{S}^\prime_i\in\mathbb{R}^{3P\times 1}.
\end{equation}

This indicates that any shape can be derived through a linear combination of other shapes. In this form, it is found that the identity matrix $\mathbf{I}$ is a trivial case for coefficient matrix $\mathbf{C}$, \ie~shapes in the sequence are not related to each other. However, the study conducted by \cite{zhu2014complex} indicates that the shapes with close distances in the deformation sequence are present in the same subspace. This guarantees the absence of trivial cases, meaning that the deformation shape sequence obtains the \textbf{self-expressive} property.

To guarantee that the estimated shape sequence has self-expressive properties, we could rebuild the sequence by applying a coefficient matrix connected to it:
\begin{align}
    \mathbf{S} &= \mathbf{C}\mathbf{S}^\prime, \mathbf{C}=\mathcal{H}(\mathbf{S}^\prime), \\~s.t.~~\mathbf{S}^\prime&=[\mathbf{S}^\prime_1,\cdots,{\mathbf S}^\prime_F]^\intercal, \mathbf{ S}^\prime_i\in\mathbb{R}^{3P\times 1}.
\end{align}
Where the mapping $\mathcal{H}(\cdot): \mathbb{R}^{F\times 3P}\to\mathbb{R}^{F\times F}$ is referenced from the \cite{qin2023toeplitz}. This mapping is designed to encode shape temporal position and similarity in the coefficient matrix at the same time. To make the network more flexible, we here first map the shapes to the feature space before imposing this regularity:
\begin{equation}
    \label{eq:context_layer}
    \mathbf{\tilde{S}} = g_s(\mathcal{H}(g(\mathbf{S'}), \mathbf{L}, \mathbf\Theta)g(\mathbf{S'})),
\end{equation}
where $\mathbf{S'}$ is the deforming sequence output from the predictor $f_\mathbf{S}(\cdot, \Theta_\mathbf{S})$ in Eq.~\eqref{eq:single_net}, $\mathbf{\Theta}$ is the learnable parameter, $\mathbf{L}$ is a series vector representing the sequence order, \ie~$\mathbf{L}=[1,2,\cdots,F]^{\intercal}$. To make the network more flexible, we impose this constraint in the feature space. $g:\mathbb{R}^{F\times 3P}\to\mathbb{R}^{F\times D}$ maps the shape to the feature space, and $g_s:\mathbb{R}^{F\times D}\to\mathbb{R}^{F\times 3P}$ remaps the deformation sequence from the feature space back to the shape space.


From the above self-expressive properties of the deformation sequence, it is clear that the shapes in the sequence are linearly correlated with each other. Therefore, the deformation sequence matrix $\mathbf{S}$ should be in canonical coordinate described by Definition~\ref{def:canoincal} and rank defect. Therefore, we perform the canonicalization with GPA Layer and the nuclear norm loss on the shape sequence, $\| \mathbf{\hat{S}}\|_*$. The nuclear norm alone cannot point to the correct learning direction, so we use the re-projection error as a data term, and finally obtain the total loss function that oversees the entire pipeline training process as follows:
\begin{equation}
    \mathcal{L} = \alpha\| \mathbf{W} - \mathbf{\Pi}\mathbf{R}\mathbf{\tilde{S}} \|_2 + \beta\| \mathbf{\hat{S}} \|_*,
\end{equation}
where $\alpha$ and $\beta$ are hyperparameters used to the balance of loss functions. After analysing the results of our experiments, we hereby choose the loss weight as $\alpha=9,\beta=0.1$.

\subsection{Implementation Details}

The complete network can be divided into three part. In the first part, we use a single-frame predictor to obtain an intermediate sequence (we here use the predictor structure of C3dpo\cite{novotny2019c3dpo}). In the second part, the intermediate sequence obtained by the single-frame predictor is first mapped to the feature space, and then regularized by the context layer to ensure that the sequence can have the self-expressive property, and then mapped back to the shape space. Finally, the output sequence is then regularized by the GPA layer for ambiguity reduction, and the reprojection error and nuclear norm are calculated using the final output shape sequence for training supervision of the pipeline.

The network module $g$ is a one-layer linear layer, the input dimension is $3\times P$, the output dimension is 128. $g_s$ contains a gated linear unit\cite{shazeer2020glu} and a four-layer MLP. $\mathcal{H}$ contains a gated Toeplitz unit (Gtu) as \cite{qin2023toeplitz}. In addition, the GPA layer is a regularization module used in training and it does not contain any learnable parameters, so it can be left out of the computation during testing. 
To clearly present the structure of our network module, we have provided more detailed information in our \textbf{supplementary material}. The experimental settings are, GPA convergence threshold is $1e^{-8}$, and the maximum number of iterations is 100. The weight of the loss function is $\alpha=9,\beta=0.1$, the sequence length is 32, the batch is 256, and 4 Nvidia RTX 3080Ti are used for training.

\section{Experiments}

We commence this section with a thorough depiction of the dataset employed as well as the metrics. Following this, the quantitative outcomes will be presented, and ultimately, the ablation experiments will be conducted, accompanied by a performance discussion.

\begin{table*}[t]
    \tabcolsep=0.13cm
    \small  
    \renewcommand\arraystretch{1.1}

    \centering
    \label{tab:nocmu}
    \begin{tabular}{c|cc|cc|cc|cc}
         \toprule 
          \multirow{2}{*}{Methods} & \multicolumn{2}{c}{GT-H36M} & \multicolumn{2}{c|}{HR-H36M} & \multicolumn{2}{c|}{I26M} & \multicolumn{2}{c}{3DPW}\\
          \cline{2-9}
          & MPJPE & Stress & MPJPE & Stress & MPJPE & Stress & MPJPE & Stress \\
         \hline
         PRN\cite{park2020procrustean}       & 86.4  &   -  & - & - & - & - & - & -\\
         PAUL\cite{wang2021paul}             & 88.3  &   -  & - & - & - & - & - & -\\
         ITES\cite{xu2021invariant}          & 77.2  &   -  & - & - & - & - & - & -\\
         PoseDict\cite{xu2021invariant}      & 85.7  &   -  & - & - & - & - & - & -\\
         C3dpo\cite{novotny2019c3dpo}        & 95.6  & 41.5 & 110.8 & 56.3 & 9.8  & 6.2 & 77.5 & 35.4\\
         DNRSfM\cite{kong2020deep}           & 109.9 & 35.9 & 121.4 & 72.4 & 13.8 & 8.5 & 184.5 & 288.1\\
         Seq2Seq\cite{deng2022deep}                          & 79.8  & 33.8 & 98.5  & 49.6 & 8.9  & 6.1 & 109.7 & 49.2\\
          MHR\cite{zeng2022mhr}                          & 72.1  & 36.4 & 93.4  & 49.9 & 29.1 & 12.2 & 79.6 & 34.7\\
         \textBF{Ours}                       & \textBF{66.1}  & \textBF{25.9} & \textBF{84.5} & \textBF{45.0} & \textBF{8.6} & \textBF{5.7} & \textBF{74.4} & \textBF{31.8} \\
         \bottomrule
    \end{tabular}
        \caption{Experimental results on the Human3.6M datasets with ground truth 2D keypoint(Marked as GT-H36M) and HRNet detected 2D keypoints(Marked as HR-H36M), InterHand2.6M(Marked as I26M) and 3DPW dataset. We report the mean per joint position error (MPJPE) over the set of test actions. Our method achieves state-of-the-art performance on these datasets.}
    \label{tab:mpjpe}
\end{table*}

\begin{table*}[t]
    \tabcolsep=0.1cm
    \renewcommand\arraystretch{1.1}
    \centering

    \begin{tabular}{c|c|ccccccccc}
    \toprule

    & Methods      & S07   & S20   & S23   & S33   & S34   & S38   & S39   & S43   & S93     \\ 
    \hline
    \multirow{7}{*}{All}
    &CSF\cite{gotardo2011computing}& 1.231 & 1.164 & 1.238 & 1.156 & 1.165 & 1.188 & 1.172 & 1.267 & 1.117  \\
    &URN\cite{cha2019unsupervised}& 1.504 & 1.770 & 1.329 & 1.205 & 1.305 & 1.303 & 1.550 & 1.434 & 1.601  \\
    &CNS\cite{cha2019reconstruct}& 0.310 & 0.217 & 0.184 & 0.177 & 0.249 & 0.223 & 0.312 & 0.266 & 0.245   \\
    &C3DPO\cite{novotny2019c3dpo}& 0.226 & 0.235 & 0.342 & 0.357 & 0.354 & 0.391 & 0.189 & 0.351 & 0.246   \\
    &Seq2Seq\cite{deng2022deep} & \underline{0.072} & \textBF{0.122} & \textBF{0.137} & \textBF{0.158} & \textBF{0.142} & \textBF{0.093} & \underline{0.090} & \textBF{0.108} & \underline{0.129}  \\
    & MHR\cite{zeng2022mhr} & 0.384 & 0.461 & 0.495 & 0.474 & 0.479 & 0.434 & 0.430 & 0.495 & 0.478 \\
    & Ours & \textbf{0.056} & \underline{0.146} & \underline{0.168} & \underline{0.169} & \underline{0.201} & \underline{0.098} & \textbf{0.082} & \underline{0.116} & \textbf{0.122}  \\
    \hline
    \multirow{6}{*}{Unseen}
    &DNRSFM & 0.097 & 0.219 & 0.264 & 0.219 & 0.209 & 0.137 & 0.127 & 0.223 & 0.164  \\
    &PR-RRN &\textBF{0.061} & \underline{0.167} &0.249 &0.254 &0.265 & 0.108 &\textBF{0.028} & \textBF{0.080} & 0.242 \\
    &C3DPO  & 0.286 & 0.361 & 0.413 & 0.421 & 0.401 & 0.263 & 0.330 & 0.491 & 0.325  \\
    &Seq2Seq & 0.081 & \textBF{0.139} & \textBF{0.196} & \textBF{0.191} & \textBF{0.195} & \underline{0.097} & 0.089 & 0.139 & \underline{0.151}  \\ 
    & MHR      & 0.398 & 0.467 & 0.511 & 0.474 & 0.471 & 0.435 & 0.428 & 0.506 & 0.476  \\
    & Ours & \underline{0.069} & {0.188} & \underline{0.228} & \underline{0.251} & \underline{0.213} & \textbf{0.096} & \underline{0.084} & \underline{0.121} & \textbf{0.120}\\
    \bottomrule
    \end{tabular}
        \caption{Results on the long sequences of the \textbf{CMU} motion capture dataset. We follow the comparison in \cite{zeng2021pr}. Our result achieve the comparable result on this dataset with the state-of-the art methods. We additionally discovered that on certain specific datasets, the benefits displayed by our created modules were not apparent. Thus, we conducted further experiments and analysed them in the Discussion section in Appendix. }
    \label{tab:cmu}
    \vspace{-2mm}
\end{table*}




\begin{table*}[t]
  \centering
      \small
    \renewcommand\arraystretch{1.1}

  \resizebox{0.95\linewidth}{!}{
  \begin{tabular}{c|ccccccc|cc}
    \toprule
    \small
    \tabcolsep=0.05cm
    \multirow{2}{*}{Methods}& \multicolumn{7}{c|}{Settings} & \multicolumn{2}{c}{Result} \\
    \cline{2-10}
     & Series & GPA & Context & Nuclear & Mean & Canonical & PRN & MPJPE & Stress \\
    \hline
     all-one Series Vector      & & \checkmark & \checkmark & \checkmark & & & & 90.4 & 36.5  \\
     w/o GPA Layer          & & & \checkmark & \checkmark & & & & 121.1 & 61.5 \\
    w/o Context Layer      & \checkmark & \checkmark & & \checkmark & & & & 166.7 & 76.8 \\
     w/o Nuclear Norm       & \checkmark & \checkmark & \checkmark  && & & & 111.2 & 45.8 \\
    Ours Canonical & \checkmark & & \checkmark & \checkmark & & \checkmark & & 79.8 & 32.6  \\
    Ours Mean     & \checkmark & & \checkmark & \checkmark &\checkmark & & & 83.4 & 33.2 \\
    Ours PRN  & \checkmark & & \checkmark & \checkmark & & & \checkmark & 118.7  & 50.8 \\
    \textBF{Ours}               & \checkmark & \checkmark & \checkmark & \checkmark  & & & & \textBF{66.1} & \textBF{25.9} \\
    \bottomrule
  \end{tabular}}
    \caption{Ablation studies performed on the Human3.6M dataset, \textbf{all-one Series Vector} representation does not take into account the temporal order at all when calculating the coefficient matrix, and its results are poor. On the other hand, the GPA Layer, Context Layer, and low rank constraint, loss of any of the constraints affect the results. \textbf{Ours Canonical} delegates do not use GPA Layer, but use \textbf{canonical loss} from C3dpo\cite{novotny2019c3dpo}, which can be found to not work well as a replacement. In addition, if we do not use GPA Layer but design a mean shape loss to replace this layer, the result as shown in \textbf{Ours Mean} and this setting does not achieve the same performance as GPA Layer. \textbf{Ours PRN} replace GPA Layer with \textbf{PRN gradient}\cite{park2020procrustean}.The proposed modules should be used together for best outcomes.}
    \label{tab:h36m-ablation}
\end{table*}

\subsection{Dataset}
\noindent{\textbf{Human3.6M.}} This classic dataset contains a large number of human motion sequences annotated with 3D ground truth which is extracted by motion capture systems \cite{h36m2014}. Following the setup of C3dpo\cite{novotny2019c3dpo}, two variants of the dataset are used: 1. Input 2D keypoints by the projection of ground truth 3D keypoints for training and testing (Marked as \emph{GT-H36M}); 2. Input 2D keypoints detected by HRNet \cite{Sun_2019_CVPR} (Marked as \emph{HR-H36M}). We follow the protocol of \cite{kudo2018unsupervised} and evaluate the mean per-joint position error(MPJPE) and Stress\cite{novotny2019c3dpo} over 17 joints.

\noindent\textbf{InterHand2.6M.} This dataset contains a large number of highly deformed hand poses \cite{moon2020interhand2}. The hand pose sequence has different deformation characteristics from the human pose sequence, and in order to verify the generality of the NRSfM method, this dataset is introduced for experimental comparison. We evaluate the mean per-joint position error (MPJPE) and Stress over 21 joints.

\noindent\textbf{3DPW.} 3DPW\cite{vonMarcard2018} is a dataset commonly used for \textit{human pose estimation}, which is less commonly used in NRSfM, and less frequently mentioned in previous work, so we follow the evaluation protocol of \cite{kudo2018unsupervised}. We also report the mean per-joint position error and Stress over 24 joints.

\noindent\textbf{CMU MOCAP.} For a fair comparison with existing methods, we follow the dataset setup of \cite{zeng2021pr} to preprocess the CMU MOCAP datasest~\footnote{CMU Motion Capture Dataset. available at \url{http://mocap.cs.cmu.edu/}}. We follow the splitting protocol of \cite{kong2020deep}, where 1/5 action sequences are used for testing, and the rest are used for training. For evaluation, we use the same criterion as \cite{zeng2021pr} and report the reconstruction error $e_{3D}$ on shapes.

\noindent\textbf{Short Sequence.} We choose some classical short sequence datasets used in the traditional methods, the Kinect dataset\cite{varol2012constrained} has 1503 points per frame and in total  191 frames; the Rug dataset\cite{garg2013variational} has 3912 points per frame and in total 159 frames. These datasets have a very small number of frames but have much denser 2D observations.

\noindent\textbf{Evaluation Metric.} In the above introduction mentioned two major metrics for evaluation, according to the object we want to compare, we choose different metrics on different data, they are:
$\textrm{MPJPE}(\mathbf{S}_i, \mathbf{S}^*_i) = \frac{1}{P}\sum_{j=1}^{P}{\|\mathbf{S}_{ij} - \mathbf{S}^*_{ij}\|_2}$ and $
\textrm{e}_{3D}(\mathbf{S}_i, \mathbf{S}^*_i)= \frac{\|\mathbf{S}_i - \mathbf{S}^*_i\|_2}{\|\mathbf{S}^*_i\|_2}$
where the $\mathbf{S}^*_{ij}$ denote the $j$-th keypoint position of $i$-th frame ground truth 3D shape and also we follow the C3dpo~\cite{novotny2019c3dpo} to report $\textrm{Stress}(\mathbf{S}_i, \mathbf{S}^*_i)=\sum_{j<k}{\frac{\|\|\mathbf{S}_{ij}-\mathbf{S}_{ik}\|_2-\|\mathbf{S}^*_{ij}-\mathbf{S}^*_{ik}\|_2\|}{P(P-1)}}$.

\subsection{Quantitative Result}

\begin{figure}[htbp]  
  \centering            
  \subfloat[H36M and I26M]   
  {
      \label{fig:viz_dataset}\includegraphics[width=0.45\linewidth]{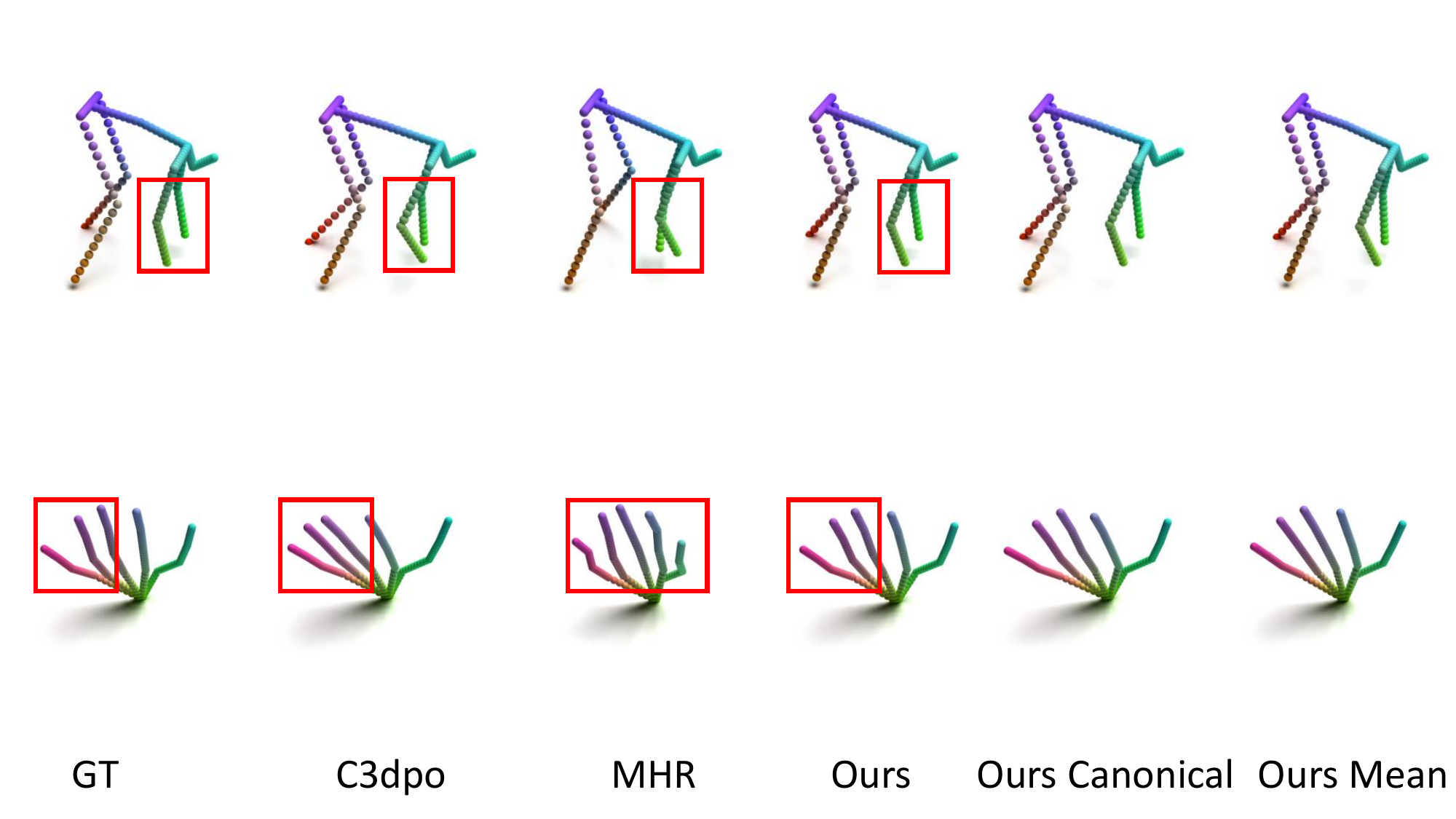}

  }
  \subfloat[Short sequence]
  {
      \label{fig:short_sequence}\includegraphics[width=0.45\linewidth]{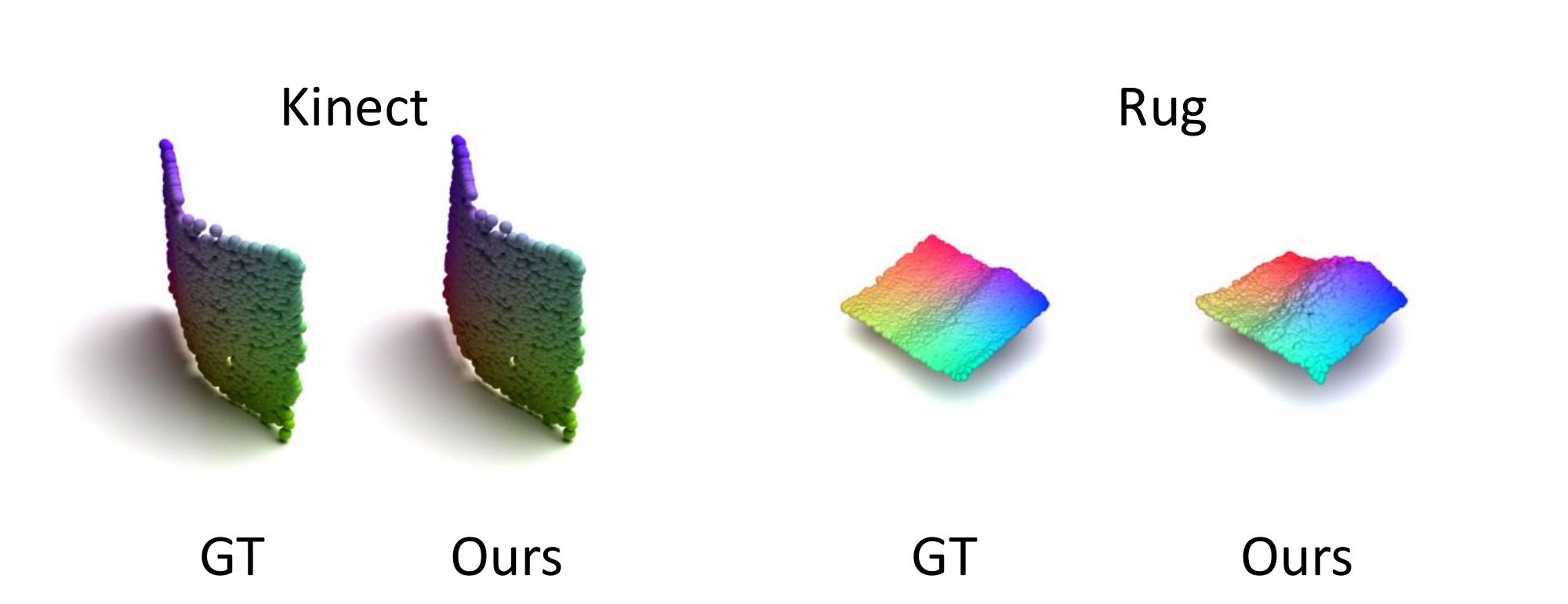}
  }
  \caption{Fig.~\ref{fig:viz_dataset} shows the visualization result on different methods and ablation on the Human3.6M dataset(first row) and InterHand2.6M (second row), more visualization are included in supplementary. Visualization in Fig.~\ref{fig:short_sequence} shows that our method also works on dense data.}    
\end{figure}

Table~\ref{tab:mpjpe} reports results on Human3.6M with ground truth keypoint (\textbf{GT-H36M}) and results on Human3.6M with detected keypoints (\textbf{HR-H36M}). We compare our result with several deep non-rigid reconstruction methods including PRN\cite{park2020procrustean}, PAUL\cite{wang2021paul}, ITES\cite{xu2021invariant}, PoseDict\cite{xu2021gtt}, C3DPO\cite{novotny2019c3dpo}, DNRSfM\cite{kong2020deep}, MHR\cite{zeng2022mhr}, Pre\cite{deng2022deep}, where we have directly quoted some experimental results from MHR. The result validates the effectiveness of our method on highly flexible shape sequences. The visualization of the result is shown in Fig.~\ref{fig:viz_dataset}. We also test this method on InterHand 2.6M as Table~\ref{tab:mpjpe}, the results validates the effectiveness of our method on dataset with category other than human.



Our method is also compared with several methods on the CMU Mocap dataset in Table~\ref{tab:cmu}. Since we use the same preprocessing method and testing strategy as PR-RRN\cite{zeng2021pr}, we directly cite the experiment results of several methods including CSF\cite{gotardo2011computing}, URN\cite{cha2019unsupervised}, CNS\cite{cha2019reconstruct} from PR-RRN, and test C3DPO\cite{novotny2019c3dpo}, MHR\cite{zeng2022mhr}, DNRSFM\cite{kong2020deep}, Seq2Seq\cite{deng2022deep} which achieves state-of-the-art (SOTA) reconstruction accuracy on CMU MOCAP, and our approach achieves comparable precision to the SOTA. 


We also test our method on the classical nonrigid dataset 3DPW. The quantitative results are reported in Table~\ref{tab:mpjpe}, showing the advantages of our method. Also, we test our method on the Kinect dataset and Rug dataset to validate dense reconstruction. The result is partially visualized in Fig.~\ref{fig:short_sequence}, and quantitative results are shown in Appendix. Although the size of these dataset are small that it is not well suited for deep methods, our method still guarantees results and is more stable compared to \cite{deng2022deep}.

We additionally discovered that on certain specific datasets and we conducted further experiments and analyzed them in the Discussion section in Appendix.



\subsection{Ablation Study}

To verify the effectiveness of our regularization module and the related regularization loss, the ablation result is reported in Table~\ref{tab:h36m-ablation}. 
The results (\textbf{Ours w/o GPA Layer, Ours w/o Context Layer}) show that using only the self-expressive constraint regularity without the alignment constraint does not achieve normal results, and vice versa. We compare the results of using the all-one Series Vector (\textbf{Our all-one Series Vector}, setting the Series Vector $\mathbf{L}={1,\cdots,1}$) and the normal vector, the results show that the correct temporal series vector in our context layer can play a role in obtaining better final shape.
In order to verify the effectiveness of the GPA layer, we replace it with loss functions of similar capability~\ie~removing the GPA layer and adding different loss functions. The results (\textbf{Ours w/o Nuclear Norm}) prove that the rank constraint for the shape sequence is effective and it can provide a positive effect for the network to find the correct learning direction. We test the Canonical loss of C3dpo (Marked as \textbf{Ours Canonical}) and the mean shape loss designed with reference to GPA respectively (\textbf{Ours Mean}) as follows:
\begin{align}  
    &\mathcal{L}_{mean} = \sum^{F}_{i=1}{\|\frac{\textrm{trace}(\mathbf{\hat R}_i)-1}{2}\|_1}, \\
    &\mathbf{\hat R}_i = \arg\min_{\mathbf R_i}\|\mathbf{R}_i\mathbf{S}_i - \mathbf{\bar{S}}\|_2,  
    \mathbf{\bar S} = \frac{1}{F}\sum^{F}_{i=1}{\mathbf S_i},
\end{align}
where $\mathbf{\hat R}_i$ is obtained by the SVD-based least square algorithm. To compare the difference between PRN and GPA Layer, we conducted \textbf{Ours PRN} by removing the GPA Layer and performing canonicalization as PRN. The results show that the GPA Layer performs better. Additionally, during the experiments, it was discovered that PRN has higher complexity and a more intricate implementation in the computation of the gradient compared to the GPA layer. 

In addition, we investigate where to supervise the reconstructed sequences, the details of which we put in the appendix due to space constraints.

\section{Conclusion}




In this paper, we revisited deep NRSfM from the perspective of canonicalization and sequence modeling. Based on our analysis, we designed two modules that can be utilized for deep NRSfM which adhere to the canonical rule. To verify the efficacy of the modules, several representative datasets are used to run the reconstruction pipeline. Experimental results confirm the validity of our approach. The findings also highlight a limitation of our method on smaller datasets. We will address this issue in our future research. 

\clearpage

\bibliography{aaai25}

\clearpage

\end{document}


\maketitle

\begin{abstract}
In this supplementary material, we provide additional discussions on the use of the context layer designed in the main text and the experiment result in the main text. Furthermore, we add more visualization of results in the form of figures and video to the experiments.

\end{abstract}

\section{Temporal information of sequence}

In traditional Non-Rigid Structure-from-Motion (NRSfM) methods, the 2D observation sequence is generally taken as a whole to solve for the corresponding 3D shape sequence. In the deep NRSfM methods, however, there is few such sequence-to-sequence modeling approach, and a single-frame lifting paradigm (\ie, solving for each 3D shape individually from a 2D frame) is usually used:
\begin{align}
   &\mathbf{S}_i = f_s(f(\mathbf{W}_i)),\mathbf{S}_i\in\mathbb{R}^{3\times P},\mathbf{W}_i\in\mathbb{R}^{3\times P},\\ 
   \nonumber &\mathbf{R} = f_R(f(\mathbf W_i)), \mathbf{R}\in\text{SO}(3),
\end{align}
where $\mathbf{S}_i$ is a 3D shape in the sequence, and similarly $\mathbf{W}_i$ is the corresponding 2D observation frame. Under the above 3D reconstruction paradigm, the following loss functions are also used as self-supervised signal sources: 
\begin{align}
    \mathcal{L} =  f_{data}(\mathbf{R},\mathbf{S},\mathbf{W}) + g(\mathbf{S}),
\end{align}
where $f_{data}(\cdot)$ is a data term, it constrains the output results that should be consistent with the modeling with the input data to produce direct supervision. The most widely data term is the reprojection error$\|\mathbf{W}-\mathbf{RS}\|$. To prevent the data terms from guiding the optimization in the wrong direction, extra regularization terms such as $g(\cdot)$ are added to further narrow down the solution space, for example, the caononicalization loss used by C3dpo\cite{novotny2019c3dpo}.

Modeling a sequence of shapes is not the same as modeling a collection of shapes. Specifically, a sequence can be viewed as a set with direction and length, so both set and temporal information (direction and length) should be taken into account when constructing the constraints. The current methods that take sequences as input objects rarely consider this issue. These methods\cite{park2020procrustean,zeng2021pr} usually take the sequence as input and subsequently construct a loss function between the members of the sequence using constraints such as smoothing or alignment. However, these constraints cannot distinguish whether the sequence has direction and distance, and the temporal information is not used.

Similarly, in \cite{deng2022deep}, 
even though the sequence order is not modeled specifically, the temporal information is added as a bias. However, such bias doesn't work too well for injecting temporal information. This is because without using downstream constraints to train such a context layer, the network is unable to recognize what exactly the bias is doing.

On the other hand, this is also determined from the computational process of the context layer. First of all, the calculation before output of the context layer can be simply described as follows:
\begin{align}
    \mathbf X_o = (\mathbf{CX} + \mathbf{\tau})\mathbf{W}_v = \mathbf{C(X+\Gamma)W}_v.
\end{align}
$\mathbf{C}$ is a weight matrix used to reconstruct the sequence, which has a relationship with the sequence as well as the temporal encoding. $\mathbf{\tau}$ is a bias that contains the temporal encoding with learnable parameters and $\mathbf{W}_v$ is an output linear layer. We found that the sequence reconstruction task can actually be done by the first term only, where the bias term would be of no direct use for the representation of sequence information. 

The bias term contains the sequence temporal encoding and learnable parameters. The loss function puts a constraint on the output of the sum of the two terms, but not specifically on the second term. This makes it uncertain what effect the second term can actually have, and it is not possible to determine whether this approach has a positive or negative effect on the outcome, making the computational process unpredictable in terms of results.

In contrast, this bias term is absent in the calculation process of this paper, and all temporal information is only included in the weight matrix, avoiding this problem without adding an additional loss function.

\begin{figure*}[htbp]
  \centering
  \includegraphics[width=1.0\linewidth]{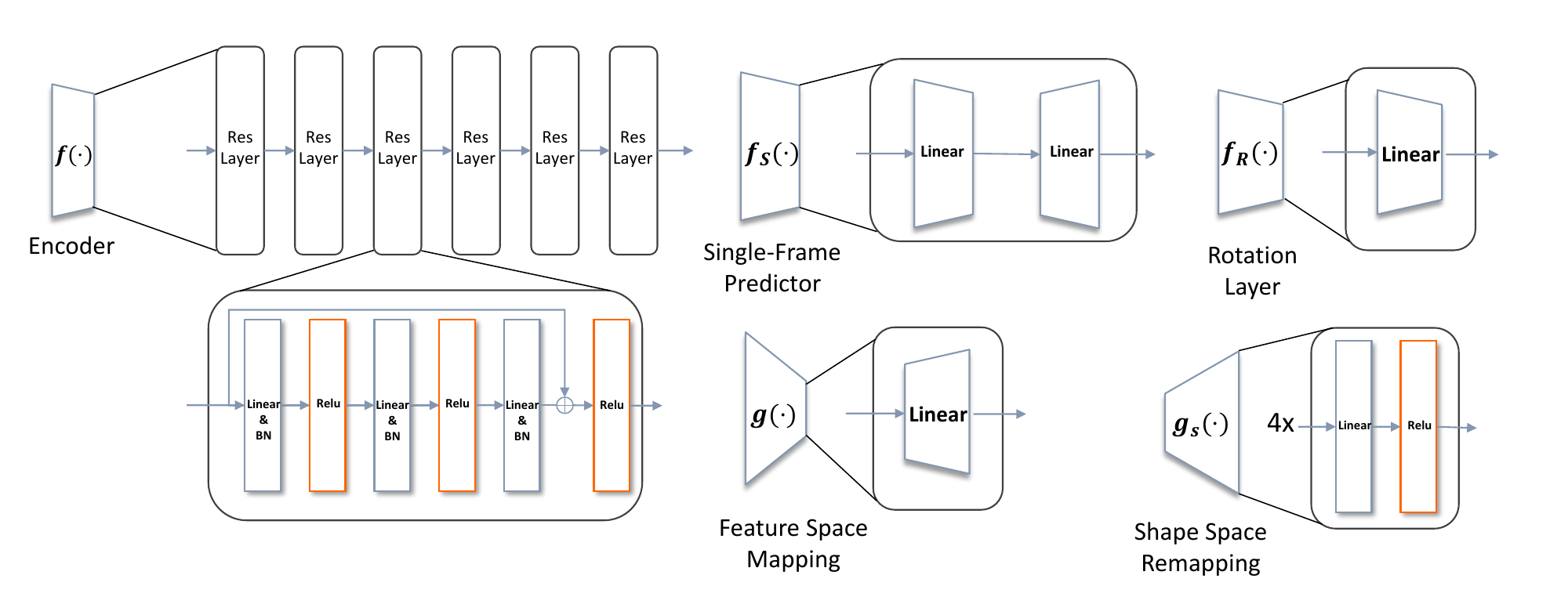}
  \caption{We have given a detailed description of the network modules in the main text, here we give a more intuitive diagram of the network modules to help the reader better understand our network structure. }
  \label{fig:module}
\end{figure*}

\section{Network architecture}
The architecture of our network modules used in main text. As shown in~\ref{fig:module}, the encoder $f(\cdot)$ contains six res-layers. The Single-Frame Predictor and the Rotation Layer utilize the decomposition network structure proposed in C3dpo~\cite{novotny2019c3dpo}. More precisely, \textit{shape predictor} comprises of two phases: the first one estimates the weights necessary for making linear combinations, while the second one computes the shapes from the weights. To ensure a computation process similar to that of the traditional factorization framework - which is linear, the structure employs purely linear layers for forming these two stage. For the same reasons, rotation layer employs a linear layer that produces a rotation vector and then uses the Rodriguez formula to convert it into a rotation matrix. The feature space mapping network module $g$ is a one-layer linear layer, the input dimension is $3\times P$, the output dimension is 128. The shape space remapping network $g_s$ contains a gated linear unit\cite{shazeer2020glu} and a four-layer MLP. $\mathcal{H}$ contains a gated Toeplitz unit (Gtu) as \cite{qin2023toeplitz}.

\begin{table}[t]
    \tabcolsep=0.255cm
    \renewcommand\arraystretch{1.1}
    \small
    \centering

    \begin{tabular}{c|ccccc}
    \toprule
    \small 
           & C3dpo & DNRSfM & MHR & Seq2Seq & Ours \\ \hline 
    Kinect & 0.191 & 0.286 & 1.15 & \textBF{0.101} & 0.166 \\
    Rug    & 0.317 & 0.351 & 0.306 & - & \textBF{0.193} \\
    \bottomrule
    \end{tabular}

    \caption{Results on the short sequences dataset. We report the $e_{3D}$ metric on this dataset. Due to the small size of this dataset, which is not well suited for data-driven deep learning methods to fully demonstrate the advantages of data-driven, the results of traditional methods are not added to the table, and are only compared in several relevant Deep NRSfM. Seq2Seq doesn't work on Rug, we mark it as '-'.}
        \label{tab:short_sequence}
\end{table}

\section{Discussion}\label{sec:discussion}

\begin{table}[t]


\small
  \tabcolsep=0.28cm
\centering
\begin{tabular}{c|ccc}

\toprule
            & Seq2Seq & \textit{Seq2Seq with GPA} & Ours  \\ \hline
CMU-S23-All & 0.137       & 0.209                & 0.168 \\ 
CMU-S34-All & 0.142       & 0.232                & 0.201 \\ 
H36M        & 79.8        & 75.6                & 66.1  \\ 
\bottomrule
\end{tabular}
\caption{We show the results of the ablation experiments, where \textit{Seq2Seq with GPA} represents the results of removing the canonical loss in the GPA Layer based on Seq2Seq. This shows the limitation of GPA Layer on datasets with shorter sequence length.}
\label{tab:conform}
\end{table}

\begin{table*}[htbp]
\caption{Performance on H36M of different pipeline structures. We only show the best result for other methods and our original strategy. }
\centering
\small
\label{tab:more_test}
\begin{tabular}{l|cccccc}
\toprule
         Weight              & 0.1                   & 0.4                  &         1           & 1.2                   & 1.5                   & 1.8                  \\ \hline
                       & MPJPE$\downarrow$                & MPJPE$\downarrow$                 & MPJPE$\downarrow$                 & MPJPE$\downarrow$                 & MPJPE$\downarrow$                 & MPJPE$\downarrow$                     \\
\hline 
C3dpo                  &      -           & -                    & 95.6                    & -                    & -                    & -                                            \\
DNRSfM                 &  -               & -                    & 109.9                    & -                    & -                    & -                                            \\
PAUL                   &  -                & -                    & 88.3                    & -                    & -                    & -                                            \\ \hline
Independent Loss                 & 97.1                 & 119.4                & 112.5                & 89.8                 & 94.8                 & 94.5                                         \\
Frozen Predictor                  &  -                & -                    & 91.3               & -                    & -                    & -   \\
Origin    & -                 & -                & \color{red}\textbf{72.5}                & -                & -                & -                                    \\
\bottomrule
\end{tabular}
\vspace{-2mm}
\end{table*}

The difference in results between Table 1 and Table 2 of main text highlights the advantages of our GPA layer on longer sequence data and the limitations on slightly shorter sequence data.  In this section, we discuss the limits and scope of different alignment regularization terms. We begin this discussion by attempting to answer why Table 1 and Table 2 of main text do not conform over comparing methods. 

The canonicalization loss used by \cite{novotny2019c3dpo} is an implicit regularization term that obtains alignment results over the entire dataset. This approach is more effective for shorter sequences or smaller datasets. However, for larger datasets, this \textbf{global alignment} becomes a bottleneck. We believe that a better approach in this case is to perform the alignment operation for each sequence individually.

However, there are some limitations to this \textbf{local alignment}. If the dataset is overall small, or if the length of each sequence in the dataset (not the input sequence) is short, then the two approaches will not show significant differences. Furthermore, if the dataset has more fragmented data, resembling a ``set'' rather than a ``sequence'', then the dictionary-based approach can also achieve better performance. To verify the idea, we conduct an analysis of the experiment results and perform ablation experiments.

\paragraph{Established Experiments.} As demonstrated in Table 1 in main text, for larger datasets like Human3.6M, Interhand2.6M, the method proposed in this paper that employs GPA Layer for local alignment substantially improves on the canonical loss method (C3dpo and Seq2Seq). Conversely, for smaller datasets like CMU Mocap shown in Table 2 in main text, the overall performance of the local alignment method seems to be similar to that of the global alignment method. If the unseen data is not incorporated into the training process, then employing the global alignment approach with canonical loss yields marginally superior outcomes.

\paragraph{Additional Experiments.} We conduct an ablation experiment to investigate the influences emanating from diverse alignment schemes. We chose two fundamental sequences from the CMU Mocap dataset and the H36M dataset as the test data. The three sets of experiments we devised are as follows: first, we tested the original Seq2Seq; second, we applied GPA Layer to replace the canonical loss of it; and finally, we introduced our proposed method for comparison. The result is shown in Table~\ref{tab:conform}. In this table, \textit{Seq2Seq with GPA} denotes that we replace the canonicalization loss in \cite{deng2022deep} with the GPA layer, and we can see that in doing so, \cite{deng2022deep} achieves comparable results on H36M, while it fails to achieve better result than the original Seq2Seq on the CMU dataset. For analogous reasons, outcomes arise for denser datasets with less data as shown in Table~\ref{tab:short_sequence}.

\paragraph{Supervision}  The reason for not directly supervising reconstructed sequence $\mathbf{S}'$ is that we want SRM to have more freedom in reconstructing. To verify this, we did related experiments at the beginning of this work. We set up different control groups: 1) We train the pipeline in two stages: first, we only train the single-frame predictor with a projection loss, and then we freeze the weights of the predictor and train the whole pipeline. 2) We train the pipeline as a whole but there is an independent projection loss to supervise $\mathbf{S}'$
. 3) Current setting. The results shows in Tabel~\ref{tab:more_test} verify our idea about.

From the results, their advantage on small datasets is also not that significant, and our results are comparable to theirs. On the other hand, it can be seen that GPA layer is not as effective on datasets with shorter sequences. This aspect serves as a constraint of the methodology suggested in this paper. Specifically, when dealing with a lesser quantity of data, the methodology puts forward in this research does not yield superior outcomes, thereby necessitating further investigation in our future work.

\section{More visualization of our result}

\begin{figure*}
  \centering
  \includegraphics[width=1.0\linewidth]{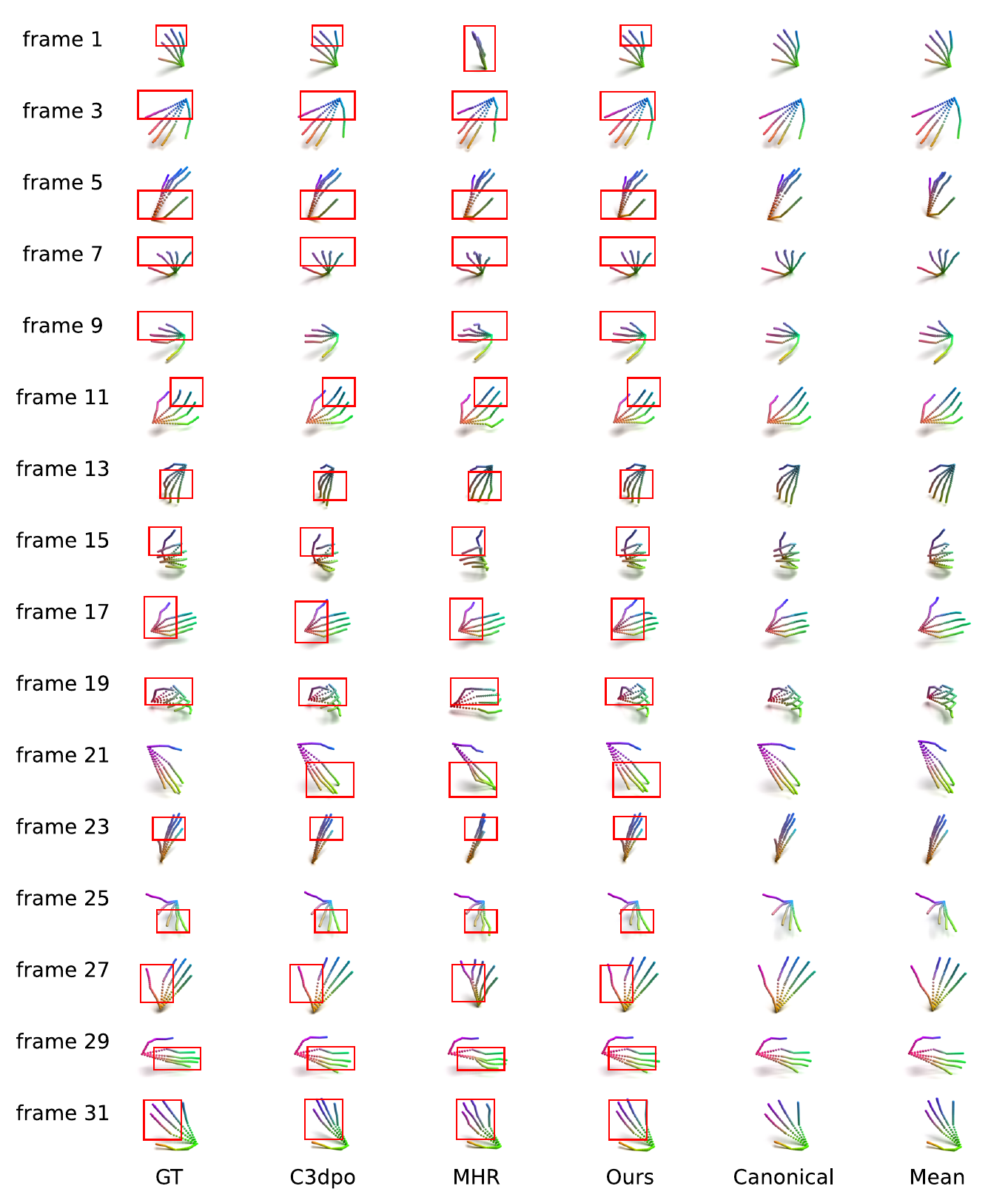}
  \caption{Visualization results of one sequence of InterHand2.6M. Different columns represent the Ground Truth and the reconstruction results of different methods. Our method shows more accurate 3D reconstruction results.}
  \label{fig:interhand}
\end{figure*}

\begin{figure*}
  \centering
  \includegraphics[width=1.0\linewidth]{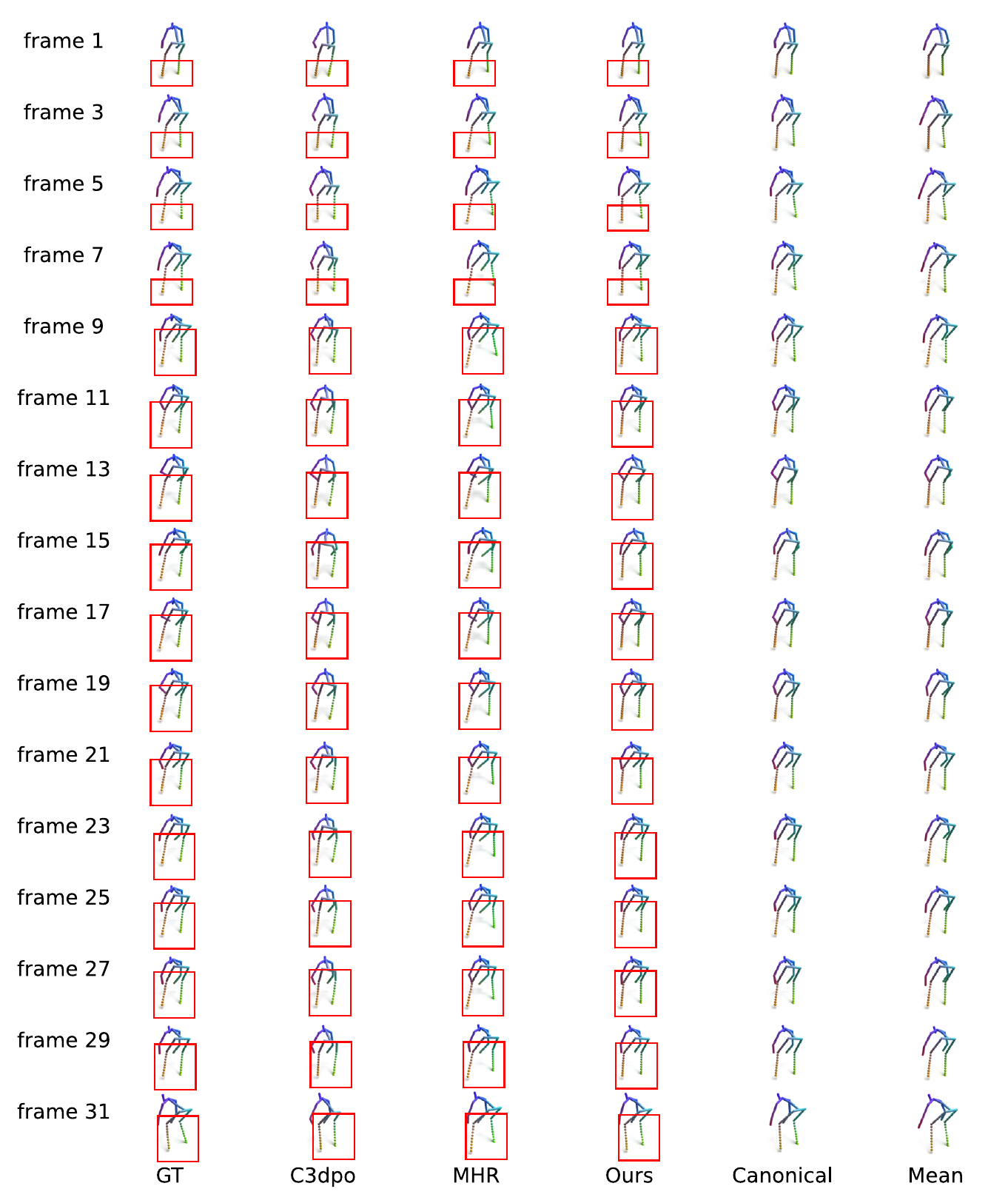}
  \caption{Visualization results of one sequence of Human3.6M. Different columns represent the Ground Truth and the reconstruction results of different methods. Our method shows more accurate 3D reconstruction results.}
  \label{fig:h36m1}
\end{figure*}

\begin{figure*}
  \centering
  \includegraphics[width=1.0\linewidth]{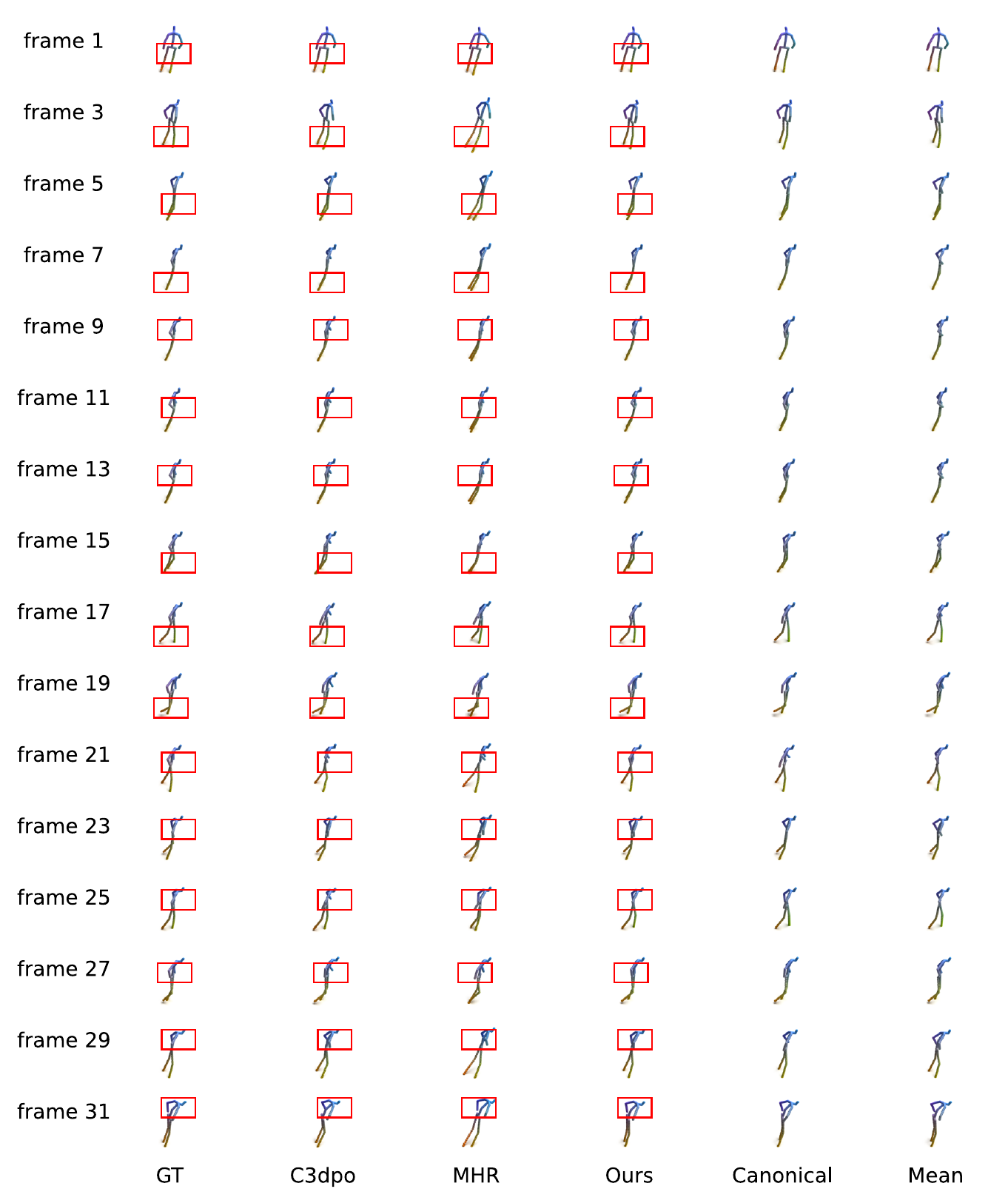}
  \caption{Visualization results of one sequence of Human3.6M. Different columns represent the Ground Truth and the reconstruction results of different methods. Our method shows more accurate 3D reconstruction results.}
  \label{fig:h36m2}
\end{figure*}

In the main text, limited by the page length, only few of the visualization of the experimental results are put, in this section, we add more visualization results to show the different methods. The experimental results under the ablation settings are shown in~\ref{fig:interhand},~\ref{fig:h36m1}, and~\ref{fig:h36m2}, where we use the \textcolor{red}{red} box to highlight the parts with relatively significant differences.

\clearpage

\bibliography{aaai25}